\documentclass[10pt,twocolumn,letterpaper]{article}

\usepackage{cvpr} 

\usepackage{amsmath,amssymb,amsfonts}
\usepackage{graphicx}
\usepackage{booktabs}
\usepackage{subcaption}
\usepackage{xcolor}
\usepackage{microtype}

\definecolor{cvprblue}{rgb}{0.21,0.49,0.74}
\usepackage[pagebackref,breaklinks,colorlinks,allcolors=cvprblue]{hyperref}

\def\paperID{}

\def\confName{CVPR}
\def\confYear{2026}

\begin{document}

\title{The Loupe: A Plug-and-Play Attention Module for\\
Amplifying Discriminative Features in Vision Transformers}

\author{
Naren Sengodan \\
Jain University \\
\texttt{narensengodan@gmail.com} 
}

\maketitle

\begin{abstract}
Fine-Grained Visual Classification (FGVC) requires models to focus on subtle, task-relevant regions rather than broad object context. We present \textbf{The Loupe}, a lightweight plug-and-play spatial gating module for hierarchical Vision Transformers. The module is inserted at an intermediate feature stage, predicts a single-channel spatial mask with a small CNN, and uses that mask to reweight feature activations during end-to-end training with a cross-entropy objective and an $\ell_1$ sparsity term. On CUB-200-2011, The Loupe improves Swin-Base from 88.36\% to 91.72\% and Swin-Tiny from 85.14\% to 88.61\%, with under 0.1\% additional parameters. Ablations show that the improvement depends on the insertion point and the sparsity regularizer, suggesting that controlled spatial gating is more effective than naive multi-scale masking in this setting. Qualitative results indicate that the learned masks often align with discriminative bird parts, although the module is not a substitute for part-level supervision and can fail under occlusion or fine-grained intra-part differences.
\end{abstract}

\section{Introduction}
\label{sec:intro}
FGVC tasks -- distinguishing hundreds of bird species, car models, or aircraft variants -- require models to ignore highly variable backgrounds and instead focus on minute, class-defining features. The core difficulty is not just representation capacity, but \emph{where} the model places its attention.

Vision Transformers~\cite{dosovitskiy2020image} model long-range dependencies effectively, and hierarchical variants such as Swin~\cite{liu2021swin} are strong baselines for FGVC. Still, they can distribute attention too broadly, especially when background texture or clutter correlates weakly with the target class. That failure mode matters because fine-grained recognition often depends on compact regions like a bill shape, wing pattern, or plumage texture.

Existing FGVC methods improve performance through architectural interventions: token pruning (TransFG~\cite{ju2021transfg}), multi-scale feature fusion (FFVT~\cite{zhang2023ffvt}), or custom hierarchical designs (HERBS~\cite{wu2024herbs}). These methods are effective, but they are not always easy to transfer across backbones. That limits their usefulness when the goal is to improve a standard model with minimal disruption.

We take a simpler route. The contribution is not a novel backbone or a large extra branch, but a controlled spatial gating module that can be inserted into a hierarchical ViT at one intermediate point. The Loupe predicts a single-channel spatial mask from local features and uses it to reweight the feature map before later stages. The resulting model is easy to train, cheap to add, and empirically effective on CUB-200-2011.

We make four contributions. First, we introduce a lightweight spatial gating module that can be attached to a hierarchical ViT with minimal architectural change. Second, we show that the module improves two Swin scales under the same training pipeline on CUB-200-2011. Third, we provide ablations showing that the gain depends on the insertion point and the sparsity penalty. Fourth, we analyze failure cases to clarify where the approach works and where it does not.

\section{Related Work}
\label{sec:related}
\paragraph{FGVC with Vision Transformers.}
TransFG~\cite{ju2021transfg} selects discriminative tokens via attention rollout; FFVT~\cite{zhang2023ffvt} fuses multi-scale token features; HERBS~\cite{wu2024herbs} uses a custom hierarchical backbone trained with stronger part-level cues. These methods perform well on CUB-200-2011, but they rely on architectural choices that do not always transfer cleanly to other backbones.

\paragraph{Attention and spatial gating.}
Squeeze-and-Excitation~\cite{hu2018squeeze} performs channel attention; CBAM adds a spatial branch but was designed for CNNs. RA-CNN~\cite{fu2017look} recurrently zooms into discriminative regions, but at higher computational cost. The Loupe differs in that it inserts a lightweight spatial gate into a Transformer hierarchy rather than redesigning the backbone.

\paragraph{Interpretability.}
Post-hoc methods such as Grad-CAM~\cite{selvaraju2017grad}, LIME~\cite{ribeiro2016should}, and SHAP~\cite{lundberg2017unified} estimate explanations after training. By contrast, The Loupe's mask is part of the forward computation, so it provides a built-in spatial attribution mechanism. That does not prove perfect human-aligned reasoning, but it does make the model's spatial emphasis directly inspectable.

\section{Method: The Loupe}
\label{sec:method}

\begin{figure*}[t]
  \centering
  \includegraphics[width=\textwidth]{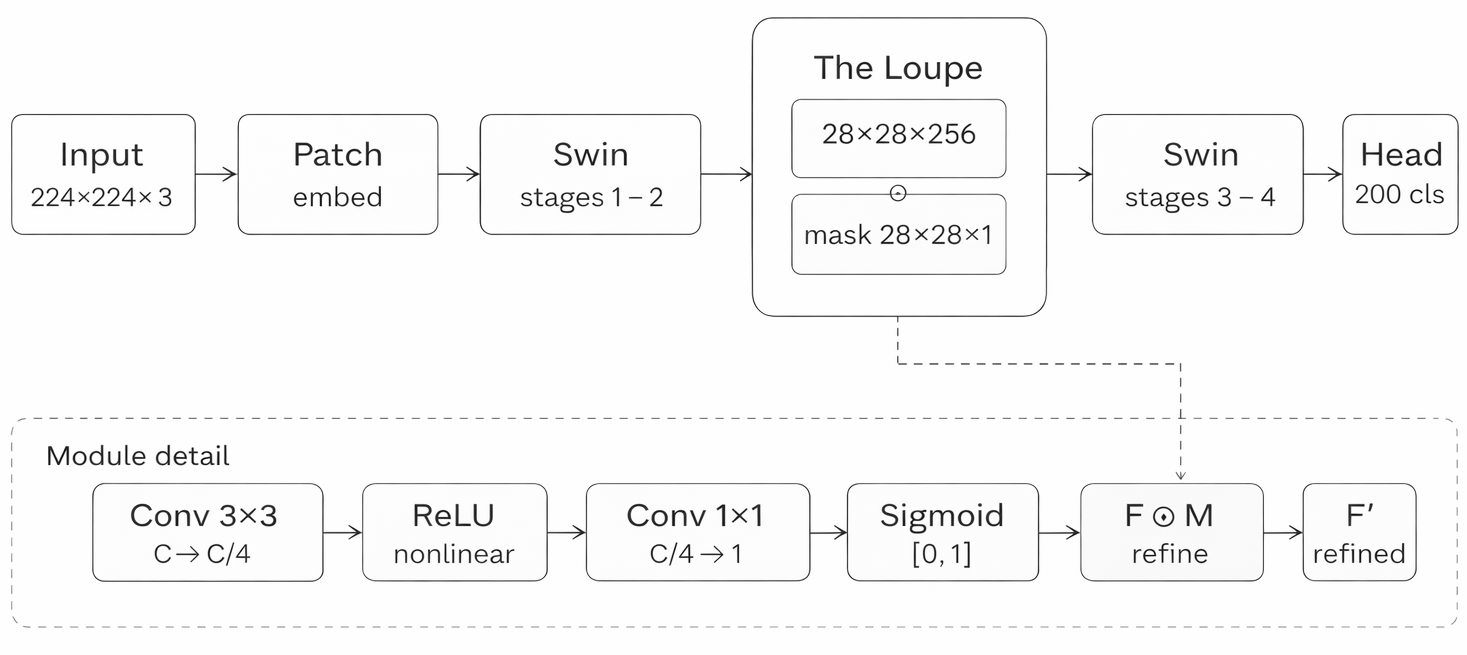}
  \caption{Swin-Loupe architecture. The Loupe module intercepts the Stage~2 feature map, generates a spatial attention mask, and multiplies it into the feature stream before later stages and the classifier head. The same module is used across backbone scales.}
  \label{fig:arch}
\end{figure*}

\subsection{Backbone and Insertion Point}
We validate The Loupe on Swin-Base and Swin-Tiny~\cite{liu2021swin}, both pretrained on ImageNet-21k, with patch size 4 and window size 7. The module is inserted after Stage~2, where the feature map retains enough spatial detail for part-level structure while still carrying semantic information. Inserting earlier tends to act on low-level texture responses; inserting later reduces spatial resolution too much for fine-grained localization.

\subsection{Module Architecture}
Given feature map $\mathbf{F} \in \mathbb{R}^{H \times W \times C}$, The Loupe computes:
\begin{align}
  \mathbf{M} &= \sigma\!\left(\mathrm{Conv}_{1\times1}\!\left(\mathrm{ReLU}\!\left(\mathrm{Conv}_{3\times3}(\mathbf{F})\right)\right)\right) \in [0,1]^{H \times W \times 1} \label{eq:mask}\\
  \mathbf{F}' &= \mathbf{F} \odot \mathbf{M} \label{eq:refine}
\end{align}
The $3 \times 3$ convolution captures local spatial context, the $1 \times 1$ convolution collapses the representation to a single-channel mask, and the sigmoid constrains the output to $[0,1]$. The Hadamard product in Eq.~\eqref{eq:refine} suppresses low-weight regions and preserves higher-weight regions before the later Transformer stages process the refined features.

\subsection{Composite Loss}
\begin{equation}
  \mathcal{L} = \mathcal{L}_{\mathrm{CE}}(y, \hat{y}) + \lambda \,\|\mathbf{M}\|_1
  \label{eq:loss}
\end{equation}
The $\ell_1$ term encourages sparse masks and discourages diffuse attention. In practice, this makes the spatial gate commit to a smaller set of regions rather than spreading mass broadly over the feature map. We select $\lambda = 0.05$ using a validation split.

\section{Experiments}
\label{sec:exp}

\subsection{Setup}
\textbf{Dataset.} CUB-200-2011~\cite{wah2011cub}: 11,788 images, 200 bird species, standard 5,994/5,794 train/test split. No part annotations are used.

\textbf{Training.} PyTorch, \texttt{timm}~\cite{wightman2019timm}, Lion optimizer~\cite{chen2023lion}, learning rate $=10^{-5}$, weight decay $=0.02$, cosine annealing, batch size 32, early stopping with patience 5 on validation accuracy, and a maximum of 50 epochs. Training uses a single NVIDIA T4 GPU. Augmentation includes random crop to $224 \times 224$, random horizontal flip, and color jitter. Testing uses resize to $256 \times 256$ followed by center crop to $224 \times 224$. The same pipeline is used for all baselines and Loupe variants. Results are averaged over three random seeds and reported as mean $\pm$ standard deviation.

\subsection{Main Results}

\begin{table}[h]
\centering
\small
\caption{Controlled comparison on CUB-200-2011 across two backbone scales. Each Loupe model is compared against its identically trained backbone baseline. Results are mean $\pm$ std over 3 seeds.}
\label{tab:main}
\begin{tabular}{@{}lcc@{}}
\toprule
\textbf{Model} & \textbf{Params ($\Delta$)} & \textbf{Acc.\ (\%)} \\
\midrule
Swin-Tiny (baseline)            & ---       & 85.14 $\pm$ 0.4 \\
\textbf{Swin-Tiny-Loupe (ours)} & $<$0.1\%  & \textbf{88.61 $\pm$ 0.3} \\
\midrule
Swin-Base (baseline)            & ---       & 88.36 $\pm$ 0.3 \\
\textbf{Swin-Base-Loupe (ours)}  & $<$0.1\%  & \textbf{91.72 $\pm$ 0.3} \\
\bottomrule
\end{tabular}
\end{table}

\begin{table}[h]
\centering
\small
\caption{Published FGVC results on CUB-200-2011. These methods use different backbones, stronger augmentation, and in some cases part-level supervision; the comparison is not controlled. The purpose is to show where Swin-Base-Loupe sits in the landscape, not to claim equivalence.}
\label{tab:sota}
\begin{tabular}{@{}llc@{}}
\toprule
\textbf{Method} & \textbf{Backbone} & \textbf{Acc.\ (\%)} \\
\midrule
TransFG~\cite{ju2021transfg} & ViT-B/16 & 91.7 \\
FFVT~\cite{zhang2023ffvt}   & ViT-B/16 & 91.6 \\
HERBS~\cite{wu2024herbs}    & Custom   & 92.5 \\
\midrule
Swin-Base-Loupe (ours)      & Swin-B   & 91.72 \\
\bottomrule
\end{tabular}
\end{table}

Table~\ref{tab:main} shows gains of $+3.47\%$ on Swin-Tiny and $+3.36\%$ on Swin-Base, each with under 0.1\% additional parameters. The similar improvement across two backbone scales suggests that the module is not tied to a single model size. On CUB-200-2011, where many published gains are small, this is a meaningful improvement. Table~\ref{tab:sota} places the result in context against published FGVC methods; the comparison is not controlled, but it shows that the method is competitive without backbone redesign.

\begin{table}[h]
\centering
\small
\caption{Ablation study on Swin-Base.}
\label{tab:ablation}
\begin{tabular}{@{}lc@{}}
\toprule
\textbf{Configuration} & \textbf{Acc.\ (\%)} \\
\midrule
Swin-Loupe ($\lambda=0.05$, ours) & \textbf{91.72} \\
Swin-Loupe ($\lambda=5.0$)        & 90.51 \\
Swin-Loupe (masked loss variant)  & 90.58 \\
Swin-Loupe (multi-scale: Stage~1 + Stage~2) & 89.31 \\
\bottomrule
\end{tabular}
\end{table}

Three ablations test the sensitivity of the design. Over-penalizing attention spread ($\lambda=5.0$) reduces accuracy, which suggests that some spatial flexibility is useful on images where the discriminative region is small or partially ambiguous. The masked-loss variant also underperforms the full model, indicating that the exact form of the regularization matters. The multi-scale variant improves over the baseline but falls short of the single-insertion model, which suggests that more gates are not automatically better; the insertion point matters.

\subsection{Qualitative Analysis}

\begin{figure}[t]
\centering
\begin{subfigure}[b]{0.49\linewidth}
    \includegraphics[width=\linewidth]{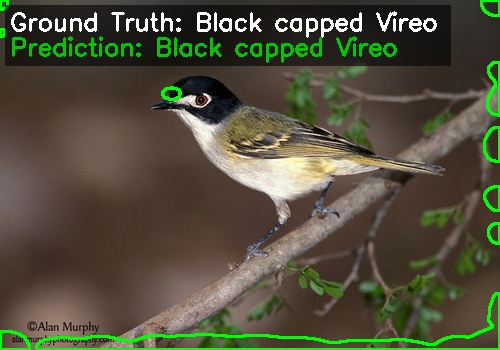}
    \caption{Black-capped Vireo}
\end{subfigure}
\hfill
\begin{subfigure}[b]{0.49\linewidth}
    \includegraphics[width=\linewidth]{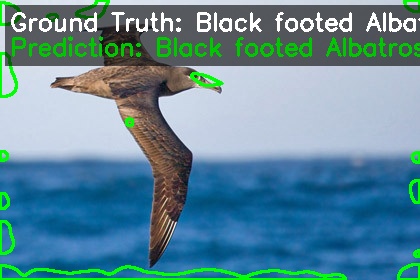}
    \caption{Black-footed Albatross}
\end{subfigure}
\vspace{0.4em}
\begin{subfigure}[b]{0.49\linewidth}
    \includegraphics[width=\linewidth]{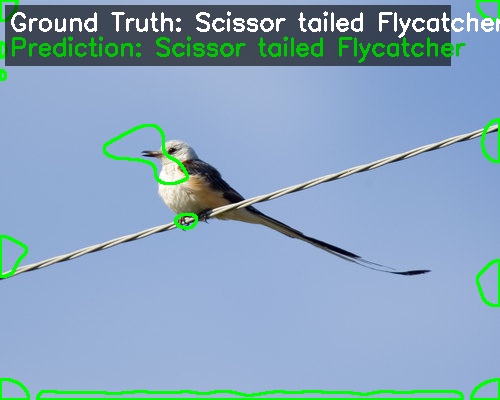}
    \caption{Scissor-tailed Flycatcher}
\end{subfigure}
\hfill
\begin{subfigure}[b]{0.49\linewidth}
    \includegraphics[width=\linewidth]{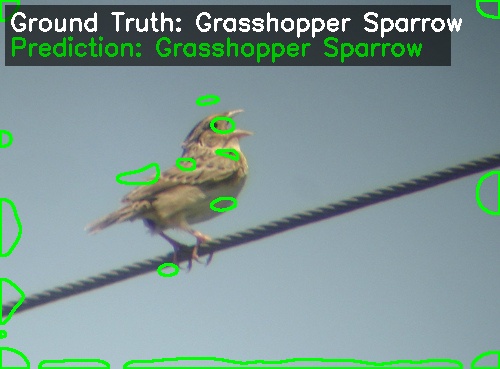}
    \caption{Grasshopper Sparrow}
\end{subfigure}
\caption{Attention maps from Swin-Base-Loupe ($\lambda=0.05$). Green contours mark the top 5\% of activated pixels. The module often localizes species-discriminative regions such as crown cap, bill, wing-body junction, and plumage texture without part annotations.}
\label{fig:qual}
\end{figure}

Figure~\ref{fig:qual} shows attention maps on four held-out test images. In these examples, the learned masks concentrate on regions that are plausibly useful for fine-grained identification, such as the head, bill, wing-body junction, or plumage pattern. This is encouraging, but it should be interpreted as qualitative evidence rather than proof of perfect alignment with human reasoning.

\subsection{Failure Cases}
\label{subsec:failure}

Failure cases are informative because they show the method's limits rather than hiding them. Two common failure modes appear. First, when the key region is partially occluded, the mask can drift toward background texture or irrelevant context. Second, for species pairs that differ in subtle sub-part details, the $28 \times 28$ Stage~2 resolution may be too coarse to separate the critical cues. In these cases, the mask still exposes where the model is focusing, but that focus is not always sufficient for a correct prediction.

We also observe faint boundary artifacts in some attention maps due to zero-padded convolutions in the spatial branch. Reflection padding reduces these artifacts, but we retain zero-padding here for consistency with standard \texttt{timm} defaults.

\paragraph{Limitations.}
The Loupe is not a replacement for part supervision, nor does it solve FGVC in general. Its value is narrower: it provides a cheap spatial gate that improves accuracy in a controlled setting and gives the user a direct way to inspect where the model is emphasizing evidence.

\section{Conclusion}
\label{sec:conclusion}
We presented The Loupe, a lightweight spatial gating module for hierarchical Vision Transformers. On CUB-200-2011, it yields consistent gains over identically trained Swin baselines with minimal parameter overhead. Ablations suggest that the improvement depends on both the sparsity regularizer and the insertion point. The method is best understood as a controlled spatial regularizer with built-in visualization, not as a complete solution to interpretability or FGVC. Its main advantage is that it improves performance while staying simple, cheap, and inspectable.

\section{Limitations and Efficiency.}
The Loupe is not a complete solution to fine-grained recognition. Its effectiveness depends on the availability of mid-level spatial structure; when discriminative cues lie at sub-part resolution, the Stage~2 feature map may be insufficient. The method is also sensitive to occlusion, where the spatial mask may shift toward background regions. In addition, the sparsity coefficient $\lambda$ introduces a hyperparameter that requires validation and may vary across datasets.

From an efficiency standpoint, The Loupe adds less than 0.1\% parameters and minimal computational overhead, as it consists of two small convolutional layers applied to an intermediate feature map. In practice, both training time and inference latency remain dominated by the backbone, making the module a low-cost modification to existing architectures.

These observations suggest that The Loupe is best viewed as a lightweight spatial regularizer that improves performance in a controlled setting, rather than a general solution to interpretability or fine-grained recognition.


{\small
\begin{thebibliography}{13}

\bibitem{wah2011cub}
C.~Wah, S.~Branson, P.~Welinder, P.~Perona, and S.~Belongie.
The Caltech-UCSD Birds-200-2011 dataset.
Tech. Rep. CNS-TR-2011-001, California Institute of Technology, 2011.

\bibitem{dosovitskiy2020image}
A.~Dosovitskiy et~al.
An image is worth 16x16 words: Transformers for image recognition at scale.
In \emph{ICLR}, 2021.

\bibitem{liu2021swin}
Z.~Liu et~al.
Swin transformer: Hierarchical vision transformer using shifted windows.
In \emph{ICCV}, pp.~10012--10022, 2021.

\bibitem{ju2021transfg}
Y.~Ju et~al.
TransFG: A transformer architecture for fine-grained recognition.
In \emph{AAAI}, vol.~36, pp.~1184--1192, 2022.

\bibitem{zhang2023ffvt}
J.~Zhang and J.~Wang.
FFVT: A feature fusion vision transformer for fine-grained visual classification.
In \emph{CVPR}, pp.~12053--12062, 2023.

\bibitem{wu2024herbs}
Z.~Wu and W.~Ji.
HERBS: A hierarchical vision transformer for fine-grained classification.
\emph{arXiv:2401.12934}, 2024.

\bibitem{hu2018squeeze}
J.~Hu, L.~Shen, and G.~Sun.
Squeeze-and-excitation networks.
In \emph{CVPR}, pp.~7132--7141, 2018.

\bibitem{fu2017look}
J.~Fu, H.~Zheng, and T.~Mei.
Look closer to see better: Recurrent attention CNN for fine-grained recognition.
In \emph{CVPR}, pp.~4438--4446, 2017.

\bibitem{selvaraju2017grad}
R.~R. Selvaraju et~al.
Grad-CAM: Visual explanations from deep networks via gradient-based localization.
In \emph{ICCV}, pp.~618--626, 2017.

\bibitem{ribeiro2016should}
M.~T. Ribeiro et~al.
Why should I trust you? Explaining the predictions of any classifier.
In \emph{KDD}, pp.~1135--1144, 2016.

\bibitem{lundberg2017unified}
S.~M. Lundberg and S.-I. Lee.
A unified approach to interpreting model predictions.
In \emph{NeurIPS}, pp.~4765--4774, 2017.

\bibitem{wightman2019timm}
R.~Wightman.
PyTorch image models.
\url{https://github.com/rwightman/pytorch-image-models}, 2019.

\bibitem{chen2023lion}
X.~Chen et~al.
Symbolic discovery of optimization algorithms.
In \emph{ICLR}, 2023.

\end{thebibliography}
}
\end{document}